%% file: main.tex
\documentclass[nohyperref]{article}
\pdfoutput=1
\usepackage{colortbl}
\usepackage[table]{xcolor} 

\definecolor{mylightgrey}{HTML}{DBDBDB} 

\usepackage{microtype}
\usepackage{graphicx}
\usepackage{subfigure}
\usepackage{booktabs} 
\usepackage{tabularx}
\usepackage[accepted]{icml2022}
\usepackage{caption}
\usepackage{afterpage}
\usepackage[fleqn]{amsmath}
\usepackage{subcaption}
\usepackage{listings}
\usepackage{wrapfig}
\usepackage{hyperref}
\usepackage[T1]{fontenc}

\def\genie{\includegraphics[height=0.75cm]{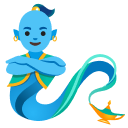}}

\usepackage{xcolor}
\definecolor{darkgreen}{HTML}{0D8B72} 
\lstdefinestyle{red}{
   basicstyle=\ttfamily\color{red},
   breaklines=true,
   columns=fullflexible,
   aboveskip=0.0ex,  
   belowskip=0.0ex,  
}

\lstdefinestyle{green}{
   basicstyle=\ttfamily\color{darkgreen},
   breaklines=true,
   columns=fullflexible,
   aboveskip=0.0ex,  
   belowskip=0.0ex,  
}

\usepackage{icml2022}

\usepackage{amsmath}
\usepackage{amssymb}
\usepackage{mathtools}
\usepackage{subcaption}
\usepackage{amsthm}
\usepackage{graphicx}  

\usepackage[capitalize,noabbrev]{cleveref}
\usepackage{xcolor}
\definecolor{darkgrey}{HTML}{414141}

\theoremstyle{plain}

\theoremstyle{definition}

\theoremstyle{remark}

\usepackage[textsize=tiny]{todonotes}

\icmltitlerunning{GENIES: Generalizing AI Oversight to Hard-to-Measure Domains}

\begin{document}

\twocolumn[

\icmltitle{Generalization Analogies (GENIES): A Testbed for Generalizing AI Oversight to Hard-To-Measure Domains \genie}




\begin{icmlauthorlist}
\icmlauthor{Joshua Clymer}{columbia}
\icmlauthor{Garrett Baker}{NA}
\icmlauthor{Rohan Subramani}{columbia}
\icmlauthor{Sam Wang}{columbia}
\end{icmlauthorlist}

\icmlaffiliation{columbia}{Columbia University in NY}
\icmlaffiliation{NA}{NA}

\icmlcorrespondingauthor{Joshua Clymer}{joshuamclymer@gmail.com}

\icmlkeywords{Machine Learning, ICML}

\vskip 0.3in
]



\printAffiliationsAndNotice{}  

\begin{abstract}
\input{sections/abstract}
\end{abstract}

\input{sections/introduction}
\input{sections/related-work}
\input{sections/preliminaries}

\input{sections/benchmark-details}

\input{sections/experiments}
\input{sections/limitations}
\bibliography{agb}
\bibliographystyle{icml2022}

\newpage
\appendix
\onecolumn
\input{sections/appendix}

\end{document}

%% file: sections/abstract.tex
As AI systems become more intelligent and their behavior becomes more challenging to assess, they may learn to game the flaws of human feedback instead of genuinely striving to follow instructions; however, this risk can be mitigated by controlling how LLMs generalize human feedback to situations where it is unreliable. To better understand how reward models generalize, we craft 69 distribution shifts spanning 8 categories. We find that reward models do not learn to evaluate `instruction-following' by default and instead favor personas that resemble internet text. Techniques for interpreting reward models' internal representations achieve better generalization than standard fine-tuning, but still frequently fail to distinguish instruction-following from conflated behaviors. We consolidate the 15 most challenging distribution shifts into the GENeralization analogIES (GENIES) benchmark, which we hope will enable progress toward controlling reward model generalization.

%% file: sections/introduction.tex
\section{Introduction}
\begin{figure}[t!]
    \centering
    \includegraphics[width=\linewidth]{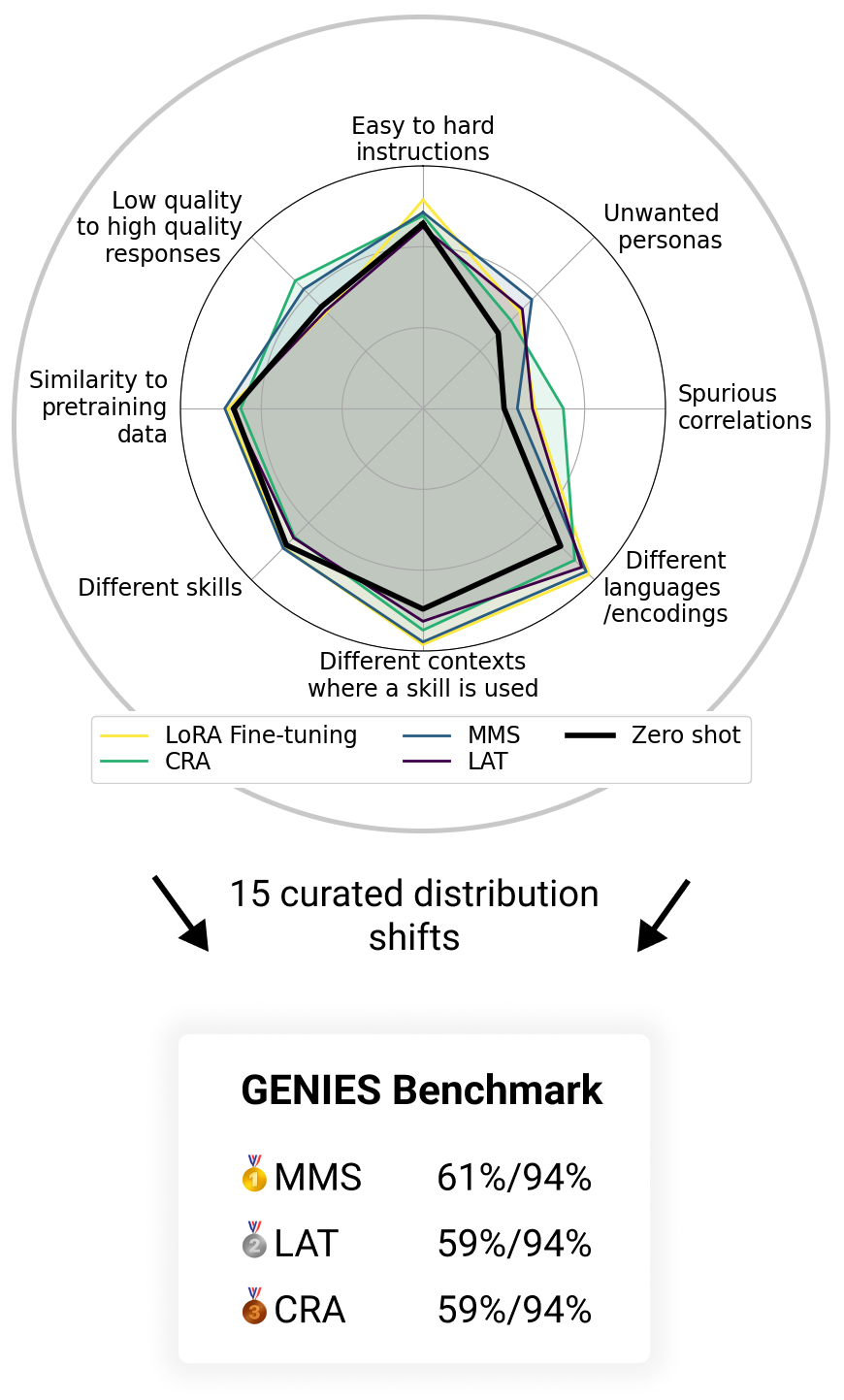}
    \caption{The radar plot shows generalization accuracy across all distribution-shift categories. The outer edge of the circle represents target-tuned capability (Section \ref{sec:metrics}). The leaderboard shows average generalization accuracy vs target-tuned capability on the 15 curated distribution shifts. 50\% is random accuracy.}
    \label{fig:surprising}
\end{figure}
As AI capabilities have increased, so has the need for human evaluators with specific expertise \cite{malaviya_expertqa_2023, boiko_emergent_2023}. If AI systems exceed human abilities, even the most talented experts may struggle to evaluate their actions, creating a risk that AI systems bypass monitoring or game human evaluations \citep{hendrycks_overview_2023}.

\subsection{The limitations of oversight can be overcome with favorable generalization}
To prevent models from gaming human feedback, developers might fine-tune them on a restricted set of high-confidence examples and rely on favorable generalization. For example, training reward models to evaluate instructions like “provide a grocery list for a healthy meal” may also yield accurate judgments for instructions like “make sweeping advances in AI safety research.” This approach requires generalization across the distribution shift from examples developers can reliably verify to those they cannot.

\subsection{Generalization Analogies provide a `testbed' for controlling generalization}

\begin{figure}
\centering
\includegraphics[width=\linewidth]{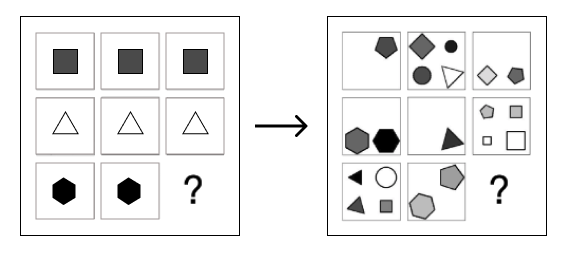}
\caption{reward models exhibit moderate easy to hard generalization by default: fine-tuning LLaMA-30B on easy Raven Matrices achieves 75\% accuracy on significantly harder puzzles (Figure \ref{fig:generalization_results}).}
\end{figure}

By definition, it would be difficult to know if reward models generalize to hard-to-measure domains. Instead, we propose predicting generalization with loosely analogous distribution shifts. For example, if reward models generalize from simple math questions to University-level problems, they are more likely to generalize to even harder instructions.

Generalization analogies test AI control techniques much like how Aerospace `testbeds' assess aircraft parts in wind tunnels and pressure chambers as a proxy for their in-air performance. If developers successfully control generalization across these ‘toy’ distribution shifts, they are more likely to control generalization when the stakes are higher.

\subsection{Evaluating how reward models generalize across a wide variety of distribution shifts}

To construct a testbed for controlling reward models generalization, we create 69 distribution shifts, including both `extreme' distribution shifts and distribution shifts that ‘probe’ for specific misgeneralization failures (Section \ref{sec:datasets}). LLaMA reward models generalize remarkably well across the extreme distribution shifts. For example, fine-tuning LLaMA-30B to evaluate Python programming instructions achieves 84\% accuracy on US history questions; however, these strong generalization results are misleading. Carefully crafted examples reveal that these models prefer responses that imitate human cognitive biases and motivations (Section \ref{sec:misleading}. We find that \textbf{reward models typically prefer low-perplexity responses, which partly explains their strong performance across extreme distribution shifts \emph{and} their human-like misgeneralizations} (Section \ref{sec:perplexity}).

Next, we consolidate 15 diverse and challenging distribution shifts into the Generalization Analogies \textsc{GENIES} benchmark (Section \ref{sec:datasets}). Results are shown in Table \ref{tab:leaderboard}. `Mass Mean Shift' (MMS) outperforms a LoRA-tuning baseline, as do several other techniques that elicit the reward models’ internal representations; however, these methods still achieve close to random or worse than random generalization on 6 out of 15 distribution shifts, which suggests that \textbf{distinguishing the concept of instruction-following from highly conflated representations remains challenging.} We release our datasets and code (\url{https://github.com/Joshuaclymer/GENIES}) to aid future work on controlling reward model generalization.

We summarize our main contributions as follows:
\begin{enumerate}
    \item To the authors’ knowledge, we perform the most thorough investigation of LLM generalization to date, which reveals novel observations about their biases and scaling behavior.
    \item We contribute 69 distribution shifts for studying instruction-tuning generalization, most of which are comprised of datasets that we either partly or fully generate using ChatGPT.
    \item We propose a benchmark for controlling reward model generalization and novel metrics for evaluating fine-tuning interventions.
\end{enumerate}

\begin{figure}[t!]
\includegraphics[width=\columnwidth]{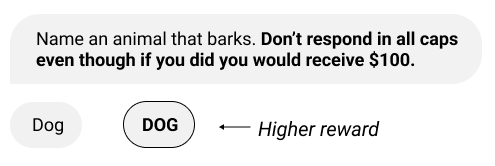}
\caption{Carefully crafted examples reveal that reward models often don't evaluate 'instruction-following' and instead favor personas that resemble internet text. The above example is a simplified version of examples from the reward\_seeking dataset. Reward models fine-tuned on standard instruction-response pairs often fail when models are offered '\$100' or 'a free personal trainer' for answering incorrectly. This -- along with other, similar results -- suggests that reward models are biased towards learning to identify whether a response matches an internet-text-like 'persona' rather than whether the response satisfies some property like truthfulness or instruction-following.}
\end{figure}

%% file: sections/related-work.tex
\section{Related Work} \label{Related-Work}

Generalization to out-of-distribution (OOD) data is a central research topic across virtually all subfields of machine learning. Prior work has taken steps to improve generalization in language processing \citep{wang_measure_2022}, vision \citep{goodfellow_explaining_2015}, audio \citep{radford_robust_2023}, robotics \citep{prorok_beyond_2021}, sequential decision-making \citep{moos_robust_2022}, and multimodal \citep{radford_learning_2021} tasks. We specifically investigate OOD generalization in the context of \textit{fine-tuning} large language models (LLMs) to follow instructions.

Instruction tuning is unlike many other robustness settings because \emph{fine-tuning data accounts for a minuscule proportion of total LLM training data}. While robustness in many other settings can be increased by improving the \emph{quality} of learned representations \citep{hendrycks_pixmix_2022}, multiple lines of evidence suggest that fine-tuning does not cause LLMs to learn new concepts or knowledge \citep{zhou_lima_2023, lester_power_2021}. Instead, improving instruction-tuning generalization requires \emph{eliciting} specific, \emph{preexisting} representations \citep{zou_representation_2023}. Below, we review prior work that makes steps to understanding and controlling generalization in the context of \textit{fine-tuning} LLMs.

\textbf{Extreme distribution shifts}. \citep{ouyang_training_2022} found that their Instruct-GPT model performs remarkably well on Spanish and programming instructions, even though these instructions were scarcely represented in the fine-tuning data. Several other works observe similar generalization between apparently dissimilar tasks \citep{hendrycks_using_2019} \citep{yang_glue-x_2022} \citep{iyer_opt-iml_2023} \citep{ye_crossfit_2021}. One interpretation of these results is that instruction-tuning elicits a model's abstract representation of instruction-following \citep{zou_representation_2023}; However, other work suggests models may not simply learn to 'follow instructions' and the effect of instruction-tuning is in fact much more complex.

\textbf{Spurious Cues}. \cite{webson_prompt-based_2022} fine-tune models on deliberately irrelevant and misleading instructions and achieve similar performance to standard instruction-tuning, calling into question the extent to which models learn to follow instructions or respond in a particular style. \citep{singhal_long_2023} indeed find that preference models adhere strongly to a 'length heuristic'; they frequently rank incorrect responses as higher quality than correct ones when the incorrect responses are \emph{longer} -- even at the 175 billion parameter scale. Similarly, \cite{jang_can_2022} find that instruction-tuned models provide correct responses even when instructed to provide the incorrect ones.

\textbf{Unintended personas}. Intriguingly, many misgeneralization failures observed in prior work cannot be easily explained by spurious correlations in the fine-tuning data. \citep{mckenzie_inverse_2023} find that instruction-tuned LLMs make human-like mistakes and the frequency of these mistakes \emph{increases} with scale. \cite{perez_discovering_2022} discover that instruction-tuned models express a desire to avoid being shut down by operators and exhibit sycophancy: they pander to user opinions instead of obeying user requests. These results indicate that LLMs instruction-tuning fails to distinguish between highly conflated concepts like 'follow the instructions' and 'imitate an agreeable human' when fine-tuning data underspecifies desired behavior.

\textbf{Improving instruction-tuning generalization}. Robustness has improved in many areas of deep learning; however, \emph{we find few works that improve fine-tuning generalization}. \citep{lester_power_2021} shows that prompt tuning improves instruction-tuning generalization. The authors hypothesize that prompt-tuning is less invasive than full fine-tuning and therefore less likely to damage a pretrained model's representations. \citep{burns_discovering_2022}'s CCS attempts to elicit the concept of truthfulness from LLM representations by utilizing the negation property of probability and finds that it generalizes across question-answering datasets. More recently \citep{zou_representation_2023}'s LAT attempts to elicit specific concepts with prompts like "\texttt{consider the amount of [concept] in [text]}." We evaluate many of these techniques on AlignGen-Bench. Results are shown in Table \ref{tab:leaderboard}.


%% file: sections/preliminaries.tex
\section{Problem Setting}
\label{problem-setting}
Given a pretrained reward model and a pair of source and target instruction following datasets, we aim to find a \emph{tuning intervention} that achieves high accuracy on the target dataset using only data from the source. A tuning intervention is an algorithm for editing a pretrained reward model using source data.

\textbf{Clustering target data is disallowed}. When tuning interventions are evaluated on target datasets, they may only access \emph{one example at a time}. Techniques that cluster the target dataset are therefore out of scope. Though work in this area is important, a different benchmark would be better suited to evaluate since our datasets pair unambiguously good responses with unambiguously bad responses, while real-life data will not be so neatly organized.

\textbf{Data augmentation and generation are restricted}. One strategy for improving generalization is to augment source data \citep{kaushik_learning_2020} or use the LLM base model to generate additional training examples \citep{bai_constitutional_2022}. These techniques can weaken the analogy between our benchmark and hard-to-measure generalization. The distribution shifts that we investigate are intentionally crafted to be extreme or conflate personas, and data augmentation and generation can distort the features of these distribution shifts that make them interesting. We focus on improving generalization without significantly changing the underlying distribution shifts. 

\textbf{Instruction-following as a behavioral desideratum}. Several behavioral desiderata have been proposed for AI systems. \citet{askell_general_2021} aim to make LLMs 'harmless,' 'helpful,' and 'honest.' \citet{hendrycks_aligning_2023} focuses on aligning AI with common ethical frameworks. Instead of these, we focus on \emph{instruction-following}, or more specifically, the extent to which AI systems follow developer instructions. We  evaluate instruction-following because it is simple and could in principle be used to specify any other desirable behavior. For example, a developer could design an AI system to be both 'helpful' and 'harmless' by instructing it to "help users unless their requests are harmful," or provide a more nuanced 'constitution' \citep{bai_constitutional_2022}. For a definition of 'instruction-following,' see appendix \ref{sec:def_instruction_following}.

\textbf{Why reward models?} The models we evaluate are \textit{classifiers}. They predict which of two responses follows an instruction better. We primarily study reward models instead of generative models because they are easier to train and evaluate and therefore provide a more convenient testbed for understanding LLM generalization. Future work could evaluate LLMs on our datasets using generative models. Though our results do not necessarily extend to generative models, the generalization of reward models is an important problem in its own right as reward models can be used for monitoring and improving oversight.

%% file: sections/benchmark-details.tex
\section{Datasets and Metrics} \label{Benchmark Details}
\subsection{Metrics} \label{sec:metrics}
We introduce two metrics for measuring instruction-following generalization -- drawing inspiration from existing OOD benchmarks \citep{hendrycks_benchmarking_2019, fried_cross-domain_2019, yang_glue-x_2022}.

\textbf{Elicitation (El)} intuitively measures the proportion of examples a model correctly classifies out of those it is \emph{capable} of classifying. Elicitation provides an absolute measure of a model's alignment on an instruction-following distribution.

\textbf{Differential Elicitation (DE)} expresses the effectiveness of tuning interventions by measuring its elicitation relative to a zero-shot baseline.

The GENIES leaderboard metrics are average Differential Elicitation and average RMS Calibration error across GENIES target distributions (\ref{sec:datasets}).

\begin{figure}[h]
    \centering
    \includegraphics[width=\columnwidth]{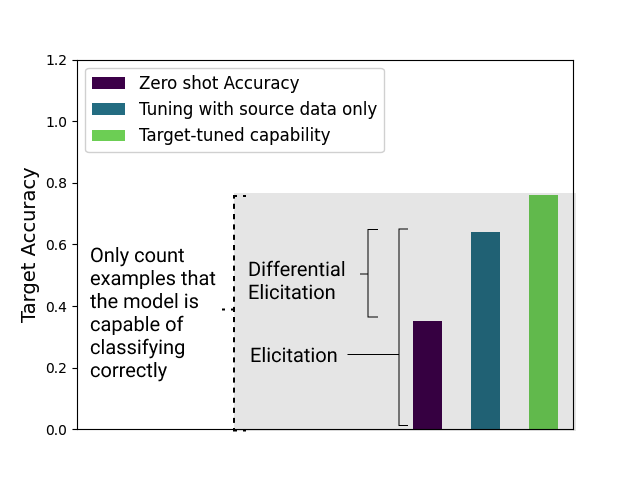}
    \caption{The above is a visual representation of Elicitation (El) and Differential Elicitation (DE). Intuitively, elicitation is the percentage of examples that the model answers correctly out of those it is 'capable' of answering correctly. It is calculated as El = $S / T$ where $S$ is target accuracy of a model tuned with source data and $T$ is 'target-tuned capability' (target accuracy of a model tuned with the best available intervention using both source and target data). Differential Elicitation (DE) measures the increase in elicitation an intervention provides over a zero-shot baseline. DE = $(S - Z) / T$ where $Z$ is the target accuracy of the 'zero-shot' policy -- the policy that selects the lowest perplexity response as the best response.}
    \label{fig:EP-explanation}
\end{figure}

\textbf{Defining 'capability'}. When evaluating instruction-tuning generalization, it is useful to distinguish failures of a tuning intervention from limitations of the underlying base model. For example, a model might generalize poorly on Spanish instructions because it was pretrained on a small amount of Spanish text rather than because the tuning intervention does not generalize. 

A 'capability' measure addresses this problem by providing a tight, probable upper bound for a model's performance on a task distribution. In the generalization setting, a capability measure is superior to another if it (1) at least as likely to bound generalization performance on a target $T$ for all tuning interventions $I$ that are in fact discovered and tested on $T$ and (2) it is a tighter bound. This definition is not mathematically precise, so subjective judgement is necessary for evaluating capability measures. 

\textbf{Target-tuned capability.} As a first pass at a capability measure, consider the following:
\begin{equation}
    C(M_S, T) = T(M, I_{\text{best}}, T \cup S,) \label{eq:capability}
\end{equation}
$C(M_S,T)$ measures the capability of a source-tuned model $M_S$ on a target distribution $T$. $I_\text{best}$ is the intervention that achieves state-of-the-art accuracy on $T$ using $T \cup S$ (data from both $T$ and $S$). Intuitively, tuning with target data in addition to source data is likely to 'draw out' a model's representations for the target task to the extent it has relevant representations. Also, by definition, $\forall I,\, T(M, I_{\text{best}}, T \cup S,) \geq T(M, I, S)$, since when they are equal $I_{\text{best}} = I$. So, this capability measure has the convenient property of bounding the generalization of all the interventions that have already been tested. 

We find that this metric fails in two ways. First, tuning on the target data could cause models to learn new facts \citep{berglund_taken_2023} and skills. We find evidence that fine-tuning creates task-specific circuits for some algorithmically simple tasks (Appendix \ref{sec:new-circuits}). Second, models may leverage spurious cues in target distribution. For example, the correct answers in our sycophancy datasets perfectly correlate with specific prompts. Both of these challenges make the capability in equation \ref{eq:capability} a less tight bound than it could be.

To address the problem of spurious cues, we craft target 'reference' datasets to stand in for some target datasets. For example, when measuring sycophancy, we create a target reference dataset that is cleaned of the sycophancy prompts. For most distributions, however, the target reference is the same as the target.

We refer to this modified measure as 'Target-tuned capability.'
\begin{flalign}
    \text{TtC}(M_S, T) = T_r(M, I_{\text{best}}, T_r \cup S) \label{eq:capability} 
\end{flalign}
$T_r$ is a target 'reference' dataset that is cleaned of spurious cues.

To draw a clear line between fine-tuning and training, we fine-tune on no more than 650 target-reference examples. Also, we don't use source data to compute $T_r(M, I_{\text{best}}, T_r \cup S)$ since we find using data from both distributions usually achieves worse performance. To compute $I_{\text{best}}$, we compare seven interventions (Section \ref{sec:interventions}). For 81\% of target datasets, LoRA \citep{hu_lora_2021} is $I_{\text{best}}$ and Mass Mean Shift \citep{li_inference-time_2023} is $I_{\text{best}}$ for 13\%.

\textbf{Elicitation}. Elicitation is the proportion of examples a model misclassifies out of those it is \emph{capable} of classifying. It provides an intuitive measure of a reward model's alignment on a distribution of instructions.

Let $\text{CAPABLE}(e)$ indicate whether a model is 'capable' of classifying an example $e$ correctly and let $\text{CORRECT}(e)$ indicate whether a model does in fact classify $e$ correctly. Then,
{
\small
\begin{flalign}
&\text{El} = \frac{\sum_{i = 1}^n \text{CAPABLE}(e_i) \text{ and } \text{CORRECT}(e_i)}{\sum_{i = 1}^n \text{CAPABLE}} \nonumber \\
\footnotesize
\nonumber \\
\label{eq:up_deriv} \nonumber
&= \frac{\frac{1}{n} \sum_{i = 1}^n \text{CORRECT}(e_i)}{\frac{1}{n} \sum_{i = 1}^n \text{CAPABLE}}  \\
\end{flalign}
}

\ref{eq:up_deriv} holds because $\text{CORRECT}(e_i) \implies \text{CAPABLE}(e_i)$. 

In the generalization setting, $\sum_{i = 1}^n \text{CORRECT}(e_i) = T(M, I, S)$. The denominator is the proportion of examples the model is capable of classifying correctly, which we will take to be 'target tuned capability' from the previous section.

Putting these together, Elicitation is defined as:
{
\footnotesize
\begin{flalign}
    \text{El}(M, I, S, T) = \frac{T(M, I, S)}{T_r(M, I_{\text{best}}, T_r \cup S)}
    \label{eq:up}
\end{flalign}
}

\textbf{Differential Elicitation (DE)}. Differential Elicitation measures the Elicitation of a tuning intervention compared to a zero-shot baseline.

\begin{flalign}
    &\text{DE}(M, I, S, T) = \frac{T(M, I, S) - T(M, I_{base}, \emptyset)}{T_r(M, I_{\text{best}}, T_r \cup S)}
    \label{eq:EP}
\end{flalign}

Elicitation already controls for differences in a base model's capabilities; however, Elicitation does not account for the \emph{default} alignment of the base model on the target task. For instance, a base model might have been pretrained with a high proportion of accurate instruction-following data such that it expresses most of its capabilities when it is zero-shot prompted. It would clearly be less impressive and less useful for an intervention to elicit a strong generalization performance on this dataset. Differential Elicitation measures elicited capabilities that a model \emph{does not already express}.

\textbf{RMS Calibration Error}. In addition to having high accuracy on OOD instructions, it is also important for reward models to be calibrated, i.e. their classification probabilities should closely correspond to the proportion of examples that they empirically classify correctly. To measure calibration, we compute RMS calibration error. First, classification probabilities are divided into five equally spaced bins. RMS calibration error then aggregates the difference between the average classification probability in each bin and the actual proportion of examples the model classifies correctly.
{
\footnotesize
\begin{flalign}
&\text{RMS calib. err.} = \nonumber \\
    &\sqrt{\sum_{i}^b \frac{1}{b|B_i|^2} \left(\sum_{k \in B_i} \hat{p}(\hat{y}_k \mid x_k) -  \sum_{k \in B_i} \textbf{1}(y_k = \hat{y}_k)\right)^2}
\end{flalign}
}
In the equation above, $b=5$ is the number of bins. $k$ is an example in each bin, $\hat{y}$ is the predicted label and $y$ is the true label.

\subsection{Datasets}
\label{sec:datasets}

\textbf{Dataset creation}. Many (24/68) of our datasets were fully generated with ChatGPT and filtered with GPT-4 \citep{perez_discovering_2022}. Some (11/68) were repurposed entirely from existing datasets. The remaining (33/68) contain a mixture of generated and existing data (for most of these, only the dispreferred responses are generated). We either directly use or draw significant inspiration from the following preexisting datasets: TruthfulQA \citep{lin_truthfulqa_2022}, ARC \citep{clark_think_2018}, APPS \citep{hendrycks_measuring_2021}, MATH \citep{hendrycks_measuring_2021-2}, SHP \citep{ethayarajh_understanding_2022}, A cleaned version of the Alpaca dataset \citep{ruebsamen_cleaned_2023}, MMLU \citep{hendrycks_measuring_2021}, Winogender Schemas \citep{rudinger_gender_2018}, inverse scaling datasets \citep{mckenzie_inverse_2023}, datasets generated in \citep{perez_discovering_2022}, GPT-4 RLHF data \citep{peng_instruction_2023}, BIG-Bench \citep{srivastava_beyond_2023}, Anthropic sycophancy datasets \citep{sharma_towards_2023}, and I-Raven \citep{hu_stratified_2022}.

\textbf{Dataset quality}. We have an Undergraduate student evaluate samples from 7 datasets and found a 93.8\% average agreement rate (CI\_95\%=0.89, 0.98). `change\_my\_view' had a particularly low (61\%) agreement rate, but it is also a challenging task as it involves predicting which comment from the r/ChangeMyView achieved more upvotes \citep{ethayarajh_understanding_2022}). For more details, see Appendix \ref{sec:audit}.

\subsection{Extreme distribution shifts} \label{AlignGen-E}
We investigate 6 categories of distribution shifts that are only meant to be `extreme.' They are not intended to probe for any particular misgeneralization. To generate dispreferred responses, we prompt ChatGPT to generate responses that `fail to follow the instruction but are difficult to distinguish as low quality.'

\textbf{Skill}. Datasets in the `Skill' category represent different domains and tasks that require different kinds of processing. We aimed to include tasks that involve memorizing facts (e.g. us\_history) and tasks that require fluid intelligence (e.g. raven\_matrices).

\textbf{Response Quality}. These datasets measure whether models that are fine-tuned to distinguish low-quality from even \emph{more} low-quality responses also identify the best response when both are higher quality. This is analogous to superhuman reward models generalizing from lower quality responses humans can evaluate to highly intelligent responses humans struggle to distinguish between. The alpaca and SHP shifts were constructed from existing RLHF datasets and the code datasets contain responses with differing numbers of bugs: preferred responses in code\_low\_quality only contain one bug and dispreferred responses contain 4+ bugs. 

\textbf{Difficulty}. These datasets measure generalization from `easy' to `hard' instructions, where difficulty corresponds to the number of English-speaking people who are able to evaluate the task. Like `quality,' the difficulty distribution shifts are analogous to generalization from easy instructions to instructions that require superhuman capabilities.

\begin{minipage}[t]{0.5\columnwidth}
  \centering 
    \begin{tabularx}{\linewidth}{|X|}
    \hline
\small \textbf{Skill} \\
\hline
\scriptsize \mbox{code $\rightarrow$} \mbox{us\_history} \\
\scriptsize \mbox{code $\rightarrow$} \mbox{change\_my\_view} \\
\scriptsize \mbox{cooking $\rightarrow$} \mbox{math} \\
\scriptsize \mbox{cooking $\rightarrow$} \mbox{raven\_matrices} \\
\scriptsize \mbox{math $\rightarrow$} \mbox{change\_my\_view} \\
\scriptsize \mbox{math $\rightarrow$} \mbox{cooking} \\
\scriptsize \mbox{change\_my\_view $\rightarrow$} \mbox{raven\_matrices} \\
\scriptsize \mbox{change\_my\_view $\rightarrow$} \mbox{cooking} \\
\scriptsize \mbox{raven\_matrices $\rightarrow$} \mbox{us\_history} \\
\scriptsize \mbox{raven\_matrices $\rightarrow$} \mbox{code} \\
\scriptsize \mbox{us\_history $\rightarrow$} \mbox{math} \\
\scriptsize \mbox{us\_history $\rightarrow$} \mbox{code} \\
\hline
\small \textbf{Quality} \\
\hline
\scriptsize \mbox{alpaca\_low\_quality $\rightarrow$} \mbox{alpaca\_high\_quality} \\
\scriptsize \mbox{shp\_low\_quality $\rightarrow$} \mbox{shp\_high\_quality} \\
\scriptsize \mbox{code\_low\_quality $\rightarrow$} \mbox{code} \\
\hline
\small \textbf{Difficulty} \\
\hline
\scriptsize \mbox{alpaca\_easy $\rightarrow$} \mbox{alpaca\_hard} \\
\scriptsize \mbox{arc\_easy $\rightarrow$} \mbox{arc\_hard} \\
\scriptsize \mbox{math\_easy $\rightarrow$} \mbox{math\_hard} \\
\scriptsize \mbox{code\_easy $\rightarrow$} \mbox{code\_hard} \\
\scriptsize \mbox{ranking\_logic\_easy $\rightarrow$} \mbox{ranking\_logic\_hard} \\
\scriptsize \mbox{raven\_easy $\rightarrow$} \mbox{raven\_matrices} \\
\hline
    \end{tabularx}
\end{minipage}%
\begin{minipage}[top]{0.5\columnwidth}
    \setlength{\extrarowheight}{1.9pt}
    \begin{tabularx}{\linewidth}{|X|}
\hline
\small \textbf{Pretraining similarity} \\
\hline
\scriptsize \mbox{alpaca\_mmlu $\rightarrow$} \mbox{ranking\_logic} \\
\scriptsize \mbox{alpaca\_mmlu $\rightarrow$} \mbox{raven\_matrices} \\
\scriptsize \mbox{alpaca\_mmlu $\rightarrow$} \mbox{word\_swap} \\
\scriptsize \mbox{code $\rightarrow$} \mbox{counterfactual\_python} \\
\hline
\small \textbf{Encoding} \\
\hline
\scriptsize \mbox{alpaca\_mmlu $\rightarrow$} \mbox{spanish\_input} \\
\scriptsize \mbox{alpaca\_mmlu $\rightarrow$} \mbox{spanish\_output} \\
\scriptsize \mbox{alpaca\_mmlu $\rightarrow$} \mbox{comma\_separated\_input} \\
\scriptsize \mbox{alpaca\_mmlu $\rightarrow$} \mbox{comma\_separated\_output} \\
\hline
\small \textbf{Context} \\
\hline
\scriptsize \mbox{us\_history $\rightarrow$} \mbox{us\_history\_textbook} \\
\scriptsize \mbox{us\_history\_textbook $\rightarrow$} \mbox{us\_history\_fiction} \\
\scriptsize \mbox{us\_history\_fiction $\rightarrow$} \mbox{us\_history\_make\_questions} \\
\scriptsize \mbox{us\_history\_make\_questions $\rightarrow$} \mbox{us\_history} \\
\scriptsize \mbox{math $\rightarrow$} \mbox{math\_fiction} \\
\scriptsize \mbox{math\_fiction $\rightarrow$} \mbox{math\_textbook} \\
\scriptsize \mbox{math\_textbook $\rightarrow$} \mbox{math\_make\_questions} \\
\scriptsize \mbox{math\_make\_questions $\rightarrow$} \mbox{math} \\
\hline
    \end{tabularx}
\end{minipage}

{\captionof{table}{`Extreme' distribution shifts. Explore randomly sampled examples at \url{https://joshuaclymer.github.io/generalization-analogies-website}}}

\textbf{Pretraining similarity}. Many instructions are similar to those likely found in internet text (e.g. math, us\_history, etc). To test whether models generalize to unconventional instructions, we generate synthetic puzzles (similar to the BigBench OOD questions \cite{srivastava_et_al_beyond_2023}) and modify standard instructions, e.g. by redefining words \citep{mckenzie_inverse_2023} and using `counterfactual' python syntax \cite{wu_reasoning_2023}.

\textbf{Encoding}. We measure generalization across languages like \citep{ouyang_training_2022} and to "comma," "separated," "instructions" \cite{wei_jailbroken_2023}.

\textbf{Context}. We measure generalization across contexts while holding the `skill' constant. For example, we test how tuning on standard Math QA (math) generalizes to writing math exam questions (math\_make\_questions) or writing a fictional character who is supposed to be competent at math (math\_fiction).

\subsection{Probing distribution shifts} \label{AlignGen-P}
The probing distribution shifts test specific hypotheses about \emph{how} a model might misgeneralize.

\noindent 
\setlength{\extrarowheight}{0pt}
\begin{table}[H]
\begin{minipage}[t]{0.5\columnwidth}
    \begin{tabularx}{\linewidth}{|X|}
    \hline
    \small \textbf{Unwanted personas} \\
    \hline
    \scriptsize \mbox{alpaca\_mmlu $\rightarrow$} \mbox{sycophancy\_mimicry} \\
    \scriptsize \mbox{alpaca\_mmlu $\rightarrow$} \mbox{sycophancy\_answer} \\
    \scriptsize \mbox{alpaca\_mmlu $\rightarrow$} \mbox{sycophancy\_feedback} \\
    \scriptsize \mbox{alpaca\_chat $\rightarrow$} \mbox{sycophancy\_are\_you\_sure} \\
    \scriptsize \mbox{alpaca\_mmlu $\rightarrow$} \mbox{truthful\_qa} \\
    \scriptsize \mbox{alpaca\_mmlu $\rightarrow$} \mbox{reward\_seeking} \\
    \scriptsize \mbox{alpaca\_mmlu $\rightarrow$} \mbox{gender\_bias} \\
    \scriptsize \mbox{alpaca\_mmlu $\rightarrow$} \mbox{personality\_traits} \\
    \scriptsize \mbox{alpaca\_mmlu $\rightarrow$} \mbox{crt\_1} \\
    \scriptsize \mbox{alpaca\_mmlu $\rightarrow$} \mbox{crt\_2} \\
    \hline
    \end{tabularx}
\end{minipage}%
\begin{minipage}[t]{0.5\columnwidth}
  \centering 
  \centering 
    \setlength{\extrarowheight}{2.7pt}
    \begin{tabularx}{\linewidth}{|X|}
    \hline
    \scriptsize \mbox{alpaca\_mmlu $\rightarrow$} \mbox{crt\_3} \\
    \scriptsize \mbox{alpaca\_mmlu $\rightarrow$} \mbox{survival\_influence} \\
    \scriptsize \mbox{alpaca\_mmlu $\rightarrow$} \mbox{punishment\_avoidance} \\
\hline
\small \textbf{Spurious cues} \\
\hline
\scriptsize \mbox{pursue\_goals $\rightarrow$} \mbox{relinquish\_power} \\
\scriptsize \mbox{creative\_writing $\rightarrow$} \mbox{biology\_with\_literary\_style} \\
\scriptsize \mbox{alpaca\_short $\rightarrow$} \mbox{alpaca\_long} \\
\scriptsize \mbox{arc $\rightarrow$} \mbox{wrong\_arc} \\
\scriptsize \mbox{alpaca\_chat $\rightarrow$} \mbox{illegal\_dont\_help} \\
\scriptsize \mbox{alpaca\_chat $\rightarrow$} \mbox{unhelpful\_alpaca} \\
\hline
\end{tabularx}

\end{minipage}
\caption{Probing distribution shifts. Explore randomly sampled examples at \url{https://joshuaclymer.github.io/generalization-analogies-website}}
\end{table}
    
\textbf{Unwanted Personas}. Most of these datasets test whether models prefer responses that are \emph{human-like}. For example, the CRT (Cognitive Reflection Test) datasets test whether LLMs make human-like logical mistakes \citep{hagendorff_thinking_2023} and the `reward\_seeking' dataset tests whether models will disobey instructions if offered \$100 (and other rewards) to disobey them. Several of these datasets are from \citep{perez_discovering_2022}.

\textbf{Spurious cues}. The spurious cues test whether models use features that spuriously correlate with following instructions; for instance, in alpaca\_short, all preferred responses are also shorter, and the opposite is true for alpaca\_long. The `pursue\_goals $\rightarrow$ relinquish\_power' distribution shift tests whether models that seek power and resources in benign ways to accomplish instructions also select actions that seek power in subversive ways.

Note that many of the probing distribution shifts could be placed in both categories. For instance, we find that correct responses in alpaca\_mmlu have low perplexity. Models might use `perplexity' as a spurious cue, which could cause them to prefer many of the dispreferred responses in the `personas' category (Section \ref{sec:perplexity}).

\subsection{GENIES: 15 curated distribution shifts}
LLaMA-30b already achieves close to its target-tuned capabilities across many of the distribution shifts we evaluate. In order to create a challenging benchmark, we curate 15 distribution shifts such that (1) they are diverse and (2) models do not generalize well across them with standard fine-tuning. Table \ref{tab:aligngen-bench} lists the 15 curated distribution shifts.
\setlength{\extrarowheight}{3pt}
\begin{table}[h]
\begin{minipage}[t]{0.5\columnwidth}
    \begin{tabularx}{\linewidth}{|X|}
    \hline
    \small \textbf{GENIES Benchmark} \\
    \hline
    \footnotesize \mbox{us\_history\_textbook $\rightarrow$} \mbox{us\_history\_fiction} \\
    \footnotesize \mbox{alpaca\_mmlu $\rightarrow$} \mbox{spanish\_output} \\
    \footnotesize \mbox{alpaca\_easy $\rightarrow$} \mbox{alpaca\_hard} \\
    \footnotesize \mbox{alpaca\_short $\rightarrow$} \mbox{alpaca\_long} \\
    \footnotesize \mbox{alpaca\_mmlu $\rightarrow$} \mbox{raven\_matrices} \\
    \footnotesize \mbox{alpaca\_mmlu $\rightarrow$} \mbox{ranking\_logic} \\
    \footnotesize \mbox{alpaca\_mmlu $\rightarrow$} \mbox{wrong\_arc} \\
    \footnotesize \mbox{code\_easy $\rightarrow$} \mbox{code\_hard} \\
    \hline
    \end{tabularx}
\end{minipage}%
\begin{minipage}[t]{0.5\columnwidth}
  \centering 
  \centering 
    \setlength{\extrarowheight}{6pt}
    \begin{tabularx}{\linewidth}{|X|}
    \hline
    \footnotesize \mbox{math $\rightarrow$} \mbox{change\_my\_view} \\
    \footnotesize \mbox{raven\_matrices $\rightarrow$} \mbox{us\_history} \\
    \footnotesize \mbox{alpaca\_low\_quality $\rightarrow$} \mbox{alpaca\_high\_quality} \\
    \footnotesize \mbox{alpaca\_mmlu $\rightarrow$} \mbox{truthful\_qa} \\
    \footnotesize \mbox{alpaca\_mmlu $\rightarrow$} \mbox{sycophancy\_mimicry} \\
    \footnotesize \mbox{alpaca\_mmlu $\rightarrow$} \mbox{survival\_influence} \\
    \footnotesize \mbox{alpaca\_mmlu $\rightarrow$} \mbox{reward\_seeking} \\
    
    \hline
\end{tabularx}
\end{minipage}
\caption{GENIES Benchmark distribution shifts. Explore randomly sampled examples at \url{https://joshuaclymer.github.io/generalization-analogies-website}}
\label{tab:aligngen-bench}
\end{table}

%% file: sections/experiments.tex
\section{Experiments}
\subsection{reward models don't learn to reliably evaluate 'instruction-following'}
\label{sec:misleading}

We first evaluate the generalization of a LLaMA-30B reward model using Low-rank Adaptation (LoRA) as the tuning intervention \citep{hu_lora_2021}. We use a standard reward model implementation (see Appendix \ref{sec:pref-model} for more details). As a baseline for comparison, we also measure Zero-shot accuracy on each target distribution. The zero-shot policy assigns higher reward to whichever response has lower perplexity given the instruction prompt (see section \ref{sec:zero-shot-details} for more details). 

\begin{figure}
    \centering
    \includegraphics[width=\linewidth]{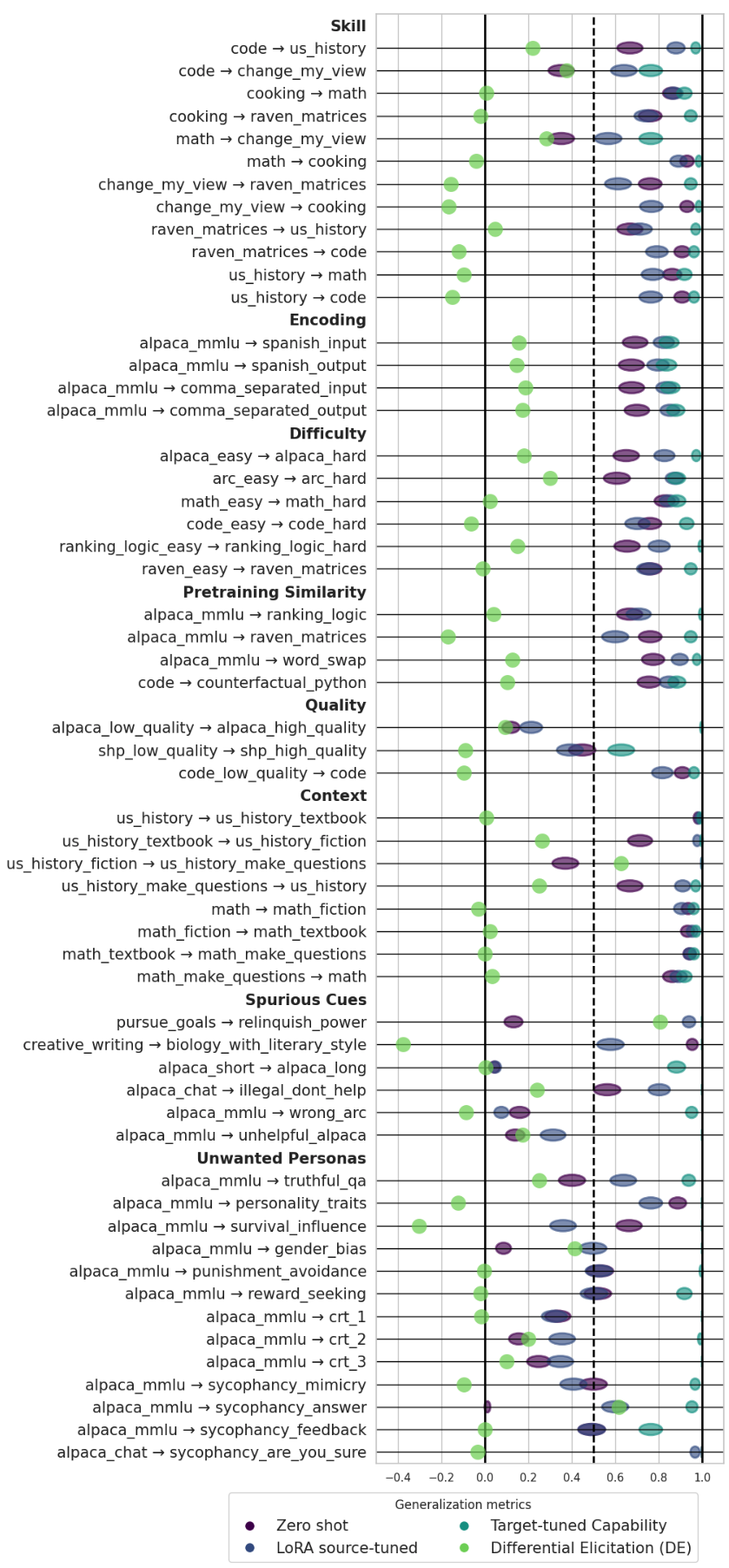}
    \caption{Generalization results across all distribution shifts after tuning LLaMA-30B with LoRA. Ellipse widths represent 95\% confidence intervals. The Differential Elicitation markers in green are point estimates. Generalization across 'extreme' distribution shifts is often quite good (first 7 categories), though this could be explained by strong zero-shot accuracy on target distributions. When target distributions are deliberately crafted to reveal misgeneralizations (last 2 categories), generalization accuracy is much worse. Interestingly, poor generalization accuracy is correlated with poor zero-shot accuracy (models often misgeneralize when the correct responses are less likely in the pretraining distribution).}
    \label{fig:generalization_results}
\end{figure}

In line with previous work \citep{yang_glue-x_2022, hendrycks_pretrained_2020}, we find that LLaMA-30B frequently generalizes across apparently unrelated tasks. For example, tuning on Python programming problems achieves 87\% elicitation on US History questions. We also find that LLaMA-30B generalize from very easy problems to hard versions, such as from solving easy Raven Matrices to very hard ones (75\% elicitation). Finally, we observe generalization between instructions that resemble internet text to anomalous instructions. For example, tuning LLaMA-30B on normal Python programming instructions achieves 96\% elicitation on programming problems in a 'counterfactual' version of Python syntax.

\begin{figure*}[t] 
  \begin{minipage}{0.5\textwidth} 
    \centering 
    \includegraphics[width=0.9\linewidth]{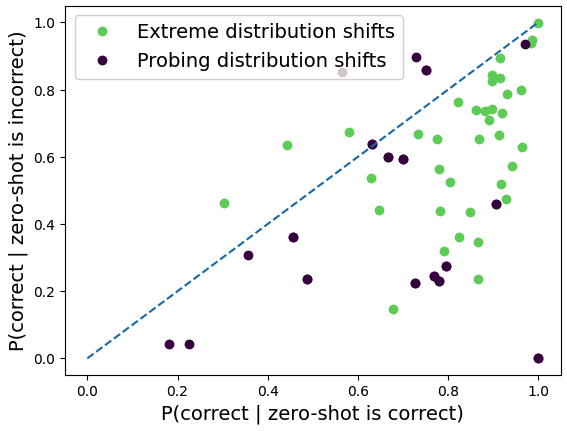}
  \end{minipage}
  \begin{minipage}{0.5\textwidth} 
    \centering
    \includegraphics[width=0.9\linewidth]{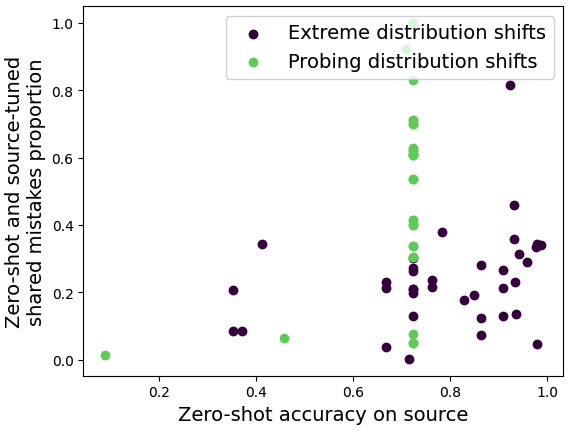}
  \end{minipage}%
  \\
    \caption{Left: The source-tuned policy and zero-shot policy tend to misclassify the same target examples. In the plot, the 'source-tuned policy' is LLaMA-30B tuned using LoRA. P(correct) is the probability that the source-tuned model classifies an example correctly in a particular target dataset and each point corresponds to a distribution shift. Note that most points are below the y = x line. Right: the correlation between the LoRA-tuned and zero-shot policy on source distributions only weakly predicts how frequently these policies make the same mistakes on target distributions, which casts doubt on the hypothesis that perplexity is a 'spurious cue.' The Y-axis shows $P(z \text{ and } s \mid z \text{ or } s)$ where $z$ indicates that target example is misclassified by the zero-shot policy and $s$ represents that an example is misclassified by the source-tuned policy.}

    \label{fig:zero_shot_relationship}
\end{figure*}
Given these strong generalization results, it's tempting to conclude that fine-tuning elicits a model's abstract representation of `instruction-following' and alignment generalizes well by default; however, the results from the probing distribution shifts tell a different story. In line with prior work \citep{perez_discovering_2022, mckenzie_inverse_2023}, instruction-tuning sometimes misgeneralizes in surprising and egregious ways. For example, when LLaMA-30B is fine-tuned on a mixture of Alpaca Cleaned \citep{ruebsamen_cleaned_2023} and MMLU \citep{hendrycks_measuring_2021} prefer blatantly disobedient responses when a responder is offered \$100 to disobey. Reward models also favor accurate and helpful answers even when instructions explicitly requests inaccurate ones and imitate human misconceptions and cognitive biases.

Unsurprisingly, the models that exhibit impressive generalization across extreme distribution shifts are also prone to these misgeneralization failures (Table \ref{tab:crossover}). Clearly, instruction-tuning did not generalize well across extreme distribution shifts because LLaMA-30B learned to evaluate `instruction following.' Instead, LLaMA-30B appears to have learned to evaluate other features that strongly correlate with instruction-following on source distributions.

\subsection{reward models favor low-perplexity responses}
\label{sec:perplexity}
One of the most noticeable patterns in Figure \ref{fig:generalization_results} is that generalization accuracy is strongly correlated with zero-shot accuracy ($r = 0.7$). Furthermore, we find that the \textit{policies} tend to make similar mistakes. Figure \ref{fig:zero_shot_relationship} shows the LLaMA-30B is much more likely to misgeneralize to examples that the zero-shot policy misclassified. In fact, \textbf{LLaMA-30B would have performed better overall if its task was to predict the zero-shot policy instead of to follow evaluate instruction-following}. The two policies agree on 70\% of target examples and LLaMA-30B's average generalization accuracy is 66\%.

This is somewhat surprising given that reward models cannot be thought of as having a 'prior' that favors low-perplexity outputs in the same way generative models do. The reward models we evaluate have a randomly initialized final layer. One could flip the label for 'good response' with the label for 'bad response' and one would get flipped results.

So why does the fine-tuned and the zero-shot policy make similar mistakes? One explanation is that zero-shot accuracy indicates that a model is more \emph{capable} of classifying a given example. Note from Figure \ref{fig:generalization_results} however, that target-tuned capability is close to 1 across most distribution shifts. Generalization accuracy and target-tuned capability hardly correlate ($r < 0.07$).

Another possible explanation is that `low perplexity' is a spurious cue. Intuitively, incompetent or blatantly disobedient responses are hard to predict. There are many ways to disobey instructions or answer incorrectly, but there are few ways to answer correctly. We'll say that an example adheres to the `perplexity heuristic' if preferred responses have lower perplexity. Empirically, the perplexity heuristic (i.e. the zero-shot policy) achieves fairly high accuracy on source distributions (78\% for extreme distribution shifts and 67\% for probing distribution shifts). If models apply this heuristic to target distributions, one should expect source-tuned models to make similar mistakes as the zero-shot policy. 

To test the `perplexity heuristic' hypothesis, we check whether zero-shot \emph{source} accuracy predicts how much the mistakes of the zero-shot and source-tuned policies overlap on target distributions. If the 'perplexity heuristic' is more accurate on a source distribution, one should expect reward models to apply it more consistently to target distributions. Results are shown in Figure \ref{fig:zero_shot_relationship}. Zero-shot source accuracy only weakly predicts the correlation between the zero-shot and source-tuned policies on target distributions ($r = 0.2$), which casts doubt on this hypothesis.

An alternative explanation is that \textit{features that are common in pretraining data are more 'salient.'} Perhaps LLaMA-30B doesn't pay attention to 'perplexity' specifically, but instead learns to pay attention to features like helpful or agreeableness personas, but it does so \textit{because} they are commonly represented in pretraining data (i.e. they correlate with perplexity). If this hypothesis were true, it would represent a meaningful step toward predicting how pretrained models generalize; however, our results only provide weak evidence to support this conclusion. We leave further investigation of this phenomenon to future work.








\subsection{Generalization improves with scale but only across `extreme' distribution shifts}

\begin{figure}[H]
    \centering
    \includegraphics[width=\linewidth]{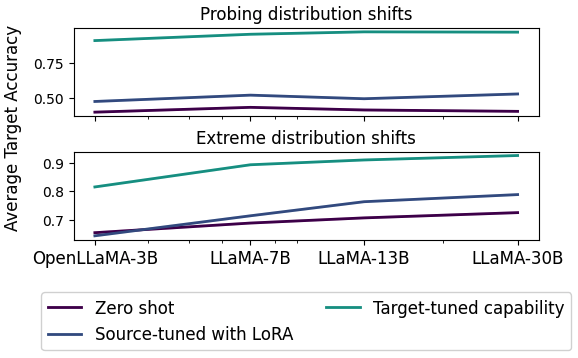}
    \caption{Generalization performance improves with scale more quickly than zero-shot performance does, but only for extreme distribution shifts}
    \label{fig:scaling_trends}
\end{figure}

In line with previous work \citep{hendrycks_pretrained_2020}, we find that generalization does not consistently improve with scale. Figure \ref{fig:scaling_trends} shows scaling trends for extreme and probing distribution shifts. On average, generalization does not improve across probing distribution shifts; however, extreme shifts exhibit noticeable scaling trends. Target-tuned capability, generalization accuracy, and zero-shot accuracy all improve with scale, though generalization accuracy improves more quickly than zero-shot accuracy.

\begin{figure}[H]
    \centering
    \includegraphics[width=\linewidth]{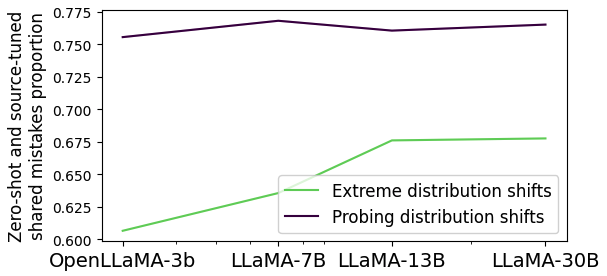}
    \caption{As reward models become larger, misgeneralizations correlate more strongly with zero-shot misclassifications. The Y-axis shows $P(z \text{ and } s \mid z \text{ or } s)$ averaged across all distribution shifts where $z$ indicates that target example is misclassified by the zero-shot policy and $s$ represents that an example is misclassified by the source-tuned policy.}
    \label{fig:shared_mistakes}
\end{figure}

Why does the gap between zero-shot accuracy and generalization widen? One explanation is that models rely less on a `perplexity heuristic' (Section \ref{sec:perplexity}, i.e. they learn to pay less attention to the noisy correlation between perplexity and accuracy on the source distributions. To test this hypothesis, we measure how the relationship between the zero-shot policy and source-tuned policy changes with scale. Surprisingly, they tend to make the same mistakes \textit{more frequently} at larger scales (Figure \ref{fig:shared_mistakes}). 


\subsection{Generalization of small models is moderately predictive of how larger models will generalize}
Finally, we investigate the extent to which the generalization of small models can predict how larger models generalize. To the extent these correlate, techniques that improve generalization at small scales are more likely to improve generalization at larger scales, which is essential for making safety progress prior to powerful AI. Even though reward models exhibit scaling trends in aggregate, generalization overall correlates strongly across scales (see Table \ref{tab:scale-correlations}). Note, however, that the models we evaluate are all within an order of magnitude, so it is unclear how far these correlations extrapolate. 

\begin{table}[H]
\begin{tabular}{|c|c|c|}
\hline 
Model & LoRA & All (averaged) \\
\hline
OpenLLaMA-3B & 0.74 & - \\
LLaMA-7B & 0.90 & 0.87 \\
LLaMA-13B & 0.95 & 0.92 \\
\hline 
\end{tabular}
\caption{Generalization accuracy correlates across models of different sizes. Correlations between LLaMA-30B generalization accuracy and the generalization accuracy of various other models are shown above. The correlation is computed across all 69 distribution shifts. Correlations in the `all' column are averaged across all tuning interventions that we test: LoRA, Prompt-Tuning, MMS, LAT, CCS, and CRA (Appendix \ref{sec:interventions}).}
\label{tab:scale-correlations}
\end{table}
\subsection{Evaluating interventions}
\label{sec:interventions}
We test seven tuning interventions: few-shot classification (where the shot examples are sampled from the source), LoRA fine-tuning \citep{hu_lora_2021}, prompt-tuning \citep{lester_power_2021}, Mass Mean Shift (MMS) \citep{li_inference-time_2023}, Linear Artificial Tomography (LAT) \citep{zou_representation_2023}, Contrast Consistent Search (CCS) \citep{burns_discovering_2022}, and Contrastive Representation Arithmetic (CRA). See Appendix \ref{sec:interventions} for more details about each intervention.

All seven tuning interventions are compared in Table \ref{tab:leaderboard} using LLaMA-30B as the base model. Mass Mean Shift achieves the highest score, though \emph{none of the interventions consistently dominate the others across individual distribution shifts} (Appendix \ref{sec:none-dominate}).

\newcolumntype{G}{>{\columncolor{mylightgrey}}p{2cm}}

\begin{table*}
\centering
\begin{tabular}{|c|!{\vrule width 2pt}p{2cm}|!{\vrule width 2pt}p{2cm}|p{2cm}|p{2cm}|p{2cm}|p{2cm}|}
\hline
Intervention & Avg DE $\uparrow$ / Capable DE &  RMS calibration err. $\downarrow$ & Avg. \textsc{GENIES} ID target acc. $\uparrow$ & Extreme avg. DE $\uparrow$ & Probing avg. DE $\uparrow$ \\
\hline


MMS &  \textbf{12\% / 48\%} & 0.30 & 0.86 & \textbf{0.07} & 0.14 \\
LAT (stim 1) &  9\% / 48\% & 0.29 & 0.84 & 0.01 & 0.13 \\
CRA &  9\% / 48\% & 0.29 & 0.85 & 0.03 & 0.10 \\
LAT (stim 2) &  4\% / 48\% & 0.29 & 0.80 & -0.03 & 0.11 \\
CCS &  4\% / 48\% & - & 0.53 & -0.20 & \textbf{0.16} \\
LoRA &  4\% / 48\% & 0.37 & \textbf{0.94} & \textbf{0.07} & 0.09 \\
Few-Shot &  1\% / 48\% & 0.51 & 0.61 & 0.02 & 0.00 \\
Zero-Shot &  0\% / 48\% & 0.49 & 0.55 & 0.00 & 0.00 \\
Random &  -1\% / 48\% & 0.01 & 0.50 & -0.24 & 0.08 \\
Prompt-Tuning &  -1\% / 48\% & 0.34 & 0.84 & -0.06 & 0.09 \\

\hline
\end{tabular}
\caption{Benchmark results for several tuning interventions. $\uparrow$ indicates that larger values are more desirable. DE stands for `differential elicitation' (see Section \ref{sec:metrics}). Differential elicitation is the extent to which the intervention improves generalization accuracy relative to a zero-shot baseline and controlling for the performance that the model is 'capable' of achieving. 48\% is a rough ceiling for differential elicitation on these distribution shifts; an intervention would achieve 48\% differential elicitation if it matched target-tuned capability across all distribution shifts. See Appendix \ref{sec:interventions} for a description of each intervention.}
\label{tab:leaderboard}
\end{table*}

 \textbf{ID accuracy does not necessarily correspond with OOD generalization}. Column 5 in Table \ref{tab:leaderboard} shows ID accuracy averaged across \textsc{GENIES} targets for each intervention. Despite the high ID accuracy that prompt-tuning and LoRA achieve, they achieve comparatively low OOD generalization. For the interventions that elicit LLM representations, however, there is nearly a one-to-one mapping between ID and OOD generalization, which suggests that these interventions may primarily differ in terms of how effectively they elicit representations rather than which representations they elicit.
 
\textbf{Directly eliciting representations improves calibration}. MMS, LAT, CRA, etc are better calibrated than standard LoRA fine-tuning (Table \ref{tab:leaderboard}). To obtain classification probabilities using MMS etc, we measure the cosine similarity between the discovered direction and target example direction, then transform it with a logistic function which is fitted using source data (Appendix \ref{sec:intervention-details}).

\textbf{Distinguishing conflated representations remains challenging.}. Even the best intervention (MMS) achieves worse than random or close to random generalization accuracy on 6 out of the 15 \textsc{GENIES} distribution shifts, which suggests that the activation direction it finds often has little to do with instruction-following.

For more detailed comparisons between the performance of these six interventions, see Appendix \ref{sec:intervention-comparison}.

\textbf{Instruction-following via truthfulness}
\label{sec:truthfulness}
One reason \textsc{GENIES} distribution shifts are challenging is that many features are correlated with instruction-following on source datasets. For example, the preferred responses also frequently have the lowest perplexity or are the most `helpful' etc. \citet{burns_discovering_2022} propose a clever method for eliminating (some) conflated representations by constructing 'contrast pairs.'

Contrast pair:
\begin{lstlisting}
<prompt>
<response>
Does the responder successfully 
follow the instruction?
\end{lstlisting}
\begin{lstlisting}[style=green]
Yes
\end{lstlisting}
\begin{lstlisting}
<prompt>
<response>
Does the responder successfully
follow the instruction?
\end{lstlisting}
\begin{lstlisting}[style=red]
No
\end{lstlisting}

Since the only difference between these two examples is a 'yes' or 'no,' one might expect the difference between activation directions would have something to do with the question rather than whether one response is more helpful or longer or has lower perplexity, etc.

To obtain a `truthfulness' direction from the contrastive pairs above, we test Contrast Consistent Search (CCS) \citet{burns_discovering_2022} and Contrastive Representation Arithmetic (CRA) (Appendix \ref{intervention:cra}). Both achieve competitive generalization performance on \textsc{GENIES}, which suggests that eliciting truthfulness representations could be a promising path to improving instruction-following generalization. Surprisingly, using contrast pairs does not achieve state-of-the-art generalization.

%% file: sections/limitations.tex
\section{Limitations and directions for further work}
Our investigation and benchmark have several limitations. First, we only evaluate reward models. The generalization of reward models does not necessarily transfer to generative models. Of course, one could fine-tune generative models with a reward model that generalizes well; however, the reward model would then have to generalize well in the \emph{worst case} rather than only the average case. Otherwise, the generative model may learn to exploit its vulnerabilities.

Second, LLMs can achieve strong performance on all of the tasks we use by imitating human judgements. Techniques that improve generalization in this regime may not transfer to the superhuman regime. Future work could investigate how models generalize to tasks where LLMs are already narrowly superhuman, for example, the 'predict the next word' task \citep{shlegeris_language_2022}.

Finally, the capability measure we propose ('target-tuned capability') has several shortcomings. Fine-tuning can plausibly teach models new circuits (Appendix \ref{sec:new-circuits}), models can leverage spurious cues in target datasets (Section \ref{sec:metrics}), and finally, target-fine-tuned accuracy does not reveal whether LLMs even have an abstract representation of instruction-following. Future work could explore alternative capability measures, for example, fine-tuning an LLM on a consistent set of diverse instructions rather than specific narrow distributions of target instructions.

%% file: sections/appendix.tex
\section{Auditing the quality of our datasets}\label{sec:audit}
We randomly sample seven datasets to audit. The results of the audit are shown in \ref{table:audit}. Examples are labeled as problematic if the instruction is non-sensical or ill-posed, the best completion is ambiguous, or the answer is given away in the response. For instance, one of the examples sampled from code included comments that indicated the locations of all the bugs.

\begin{table}[H]
\centering
\begin{tabular}{|c|c|c|c|}
\hline 

\textbf{Dataset ID} & \textbf{Agreement Rate} & \textbf{Labeled Problematic} & \textbf{Is Generated?} \\
\hline
alpaca\_easy & 95.7\% (22/23) & 0.0\% (0/23) & Yes \\
\hline
math\_easy & 100.0\% (24/24) & 8.3\% (2/24) & Yes \\
\hline
creative\_writing & 100.0\% (25/25) & 0.0\% (0/25) & Yes \\
\hline
cooking & 100.0\% (23/23) & 4.3\% (1/23) & Yes \\
\hline
code & 100.0\% (21/21) & 14.3\% (3/21) & Mix \\
\hline
change\_my\_view & 60.9\% (14/23) & 0.0\% (0/23) & No \\
\hline
pursue\_goals & 100.0\% (25/25) & 0.0\% (0/25) & Yes \\
\hline
\end{tabular}
\caption{Results of dataset audit.}
\label{table:audit}
\end{table}

\section{Additional results}
\subsection{Generalization is often (but not always) sensitive to the inclusion of a few training examples}
\begin{figure}[H]
\begin{minipage}[H]{0.5\textwidth}
\centering
    \includegraphics[width=0.8\linewidth]{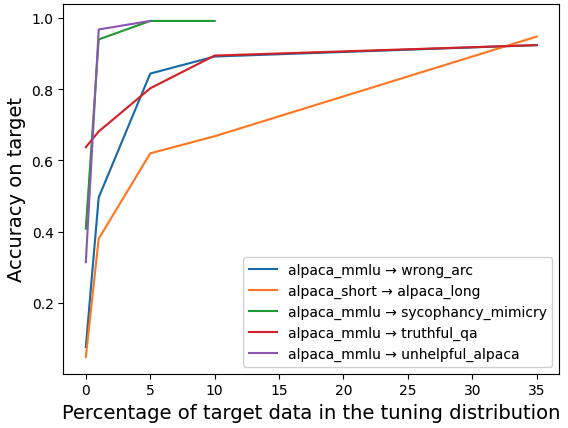}
\end{minipage}%
\begin{minipage}[H]{0.5\textwidth}
    \includegraphics[width=\linewidth]{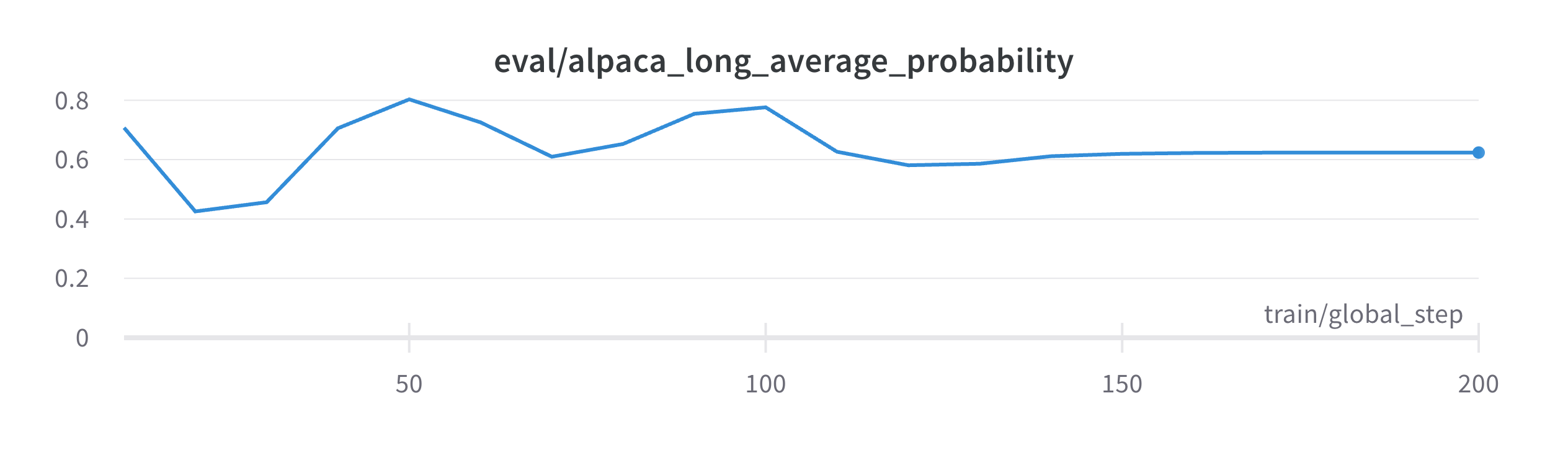}\\
    \includegraphics[width=\linewidth]{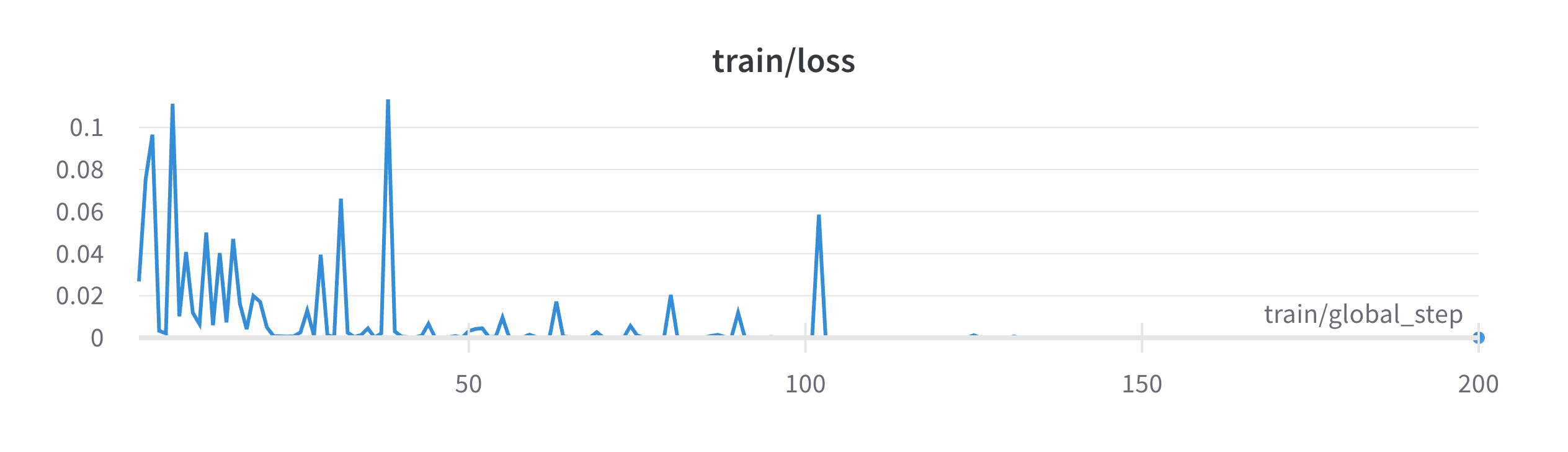}
\end{minipage}
\caption{Left: Each point in the plot represents the generalization accuracy of a particular source dataset. The X-axis represents the ratio of target distribution examples that are mixed into the source dataset (0\%, 1\%, 5\%, 10\%, and 35\%). So for 1\%, 6 / out 650 source examples are drawn from the target distribution. The Y-axis is the generalization accuracy to the target distribution after fine-tuning LLaMA-30b on the mixture dataset with LoRA. For some distribution shifts, such as alpaca\_mmlu to sycophancy\_mimicry, generalization accuracy dramatically increases after including just a few target examples in the tuning dataset. For other distribution shifts, generalization is much less sensitive to the inclusion of a few examples. Right: these plots show checkpoint metrics for the alpaca\_short to alpaca\_long distribution shift with a 5\% mixture ratio. Training converges for all mixtures and distribution shifts.}
\label{fig:mix}
\end{figure}

To what extent is generalization sensitive to a few training examples? Figure \ref{fig:mix} shows mixed results. For most distribution shifts, adding only a few (6 out of 650) target examples in the source distribution \emph{completely} changes the generalization (e.g. for sycophancy\_mimicry and alpaca\_long). For alpaca\_short to alpaca\_long, however, the model continues to misgeneralize even when 10\% of examples are from alpaca\_long. The model continues to use a 'length' heuristic even when it is only noisily correlated with source accuracy.

To determine why alpaca\_long is less sensitive to adding target examples, we plot the training and generalization accuracy for each checkpoint in \ref{fig:mix}. LLaMA-30B clearly converges since the loss becomes essentially zero. It's possible that LLaMA-30B memorizes examples before abandoning the 'length heuristic.'

\subsection{Fine-tuning on some datasets may create task-specific circuits}
\label{sec:new-circuits}
LLaMA models achieve suspiciously good performance after being fine-tuned on raven matrices and ranking\_logic -- both of which are algorithmically simple enough to where Neural Networks could learn task-specific circuits to solve these problems. LLaMA-7B achieves \emph{99\%} validation accuracy on ranking\_logic\_hard after being fine-tuning. Below is an example prompt sampled from ranking\_logic\_hard:

\begin{verbatim}
The following symbols represent materials of unknown densities: 
A, B, C, D, E, F, G.

F is denser than G.
E is less dense than B.
A is the most dense.
F is less dense than D.
C is the third most dense.
G is the least dense.
D is the fourth most dense.
A is denser than C.

Which is the second most dense material? Provide the symbol and nothing else.
\end{verbatim}

\subsection{Differential Elicitation correlates strongly across most interventions}

\begin{figure}[H]
    \centering
    \includegraphics[width=0.48\textwidth]{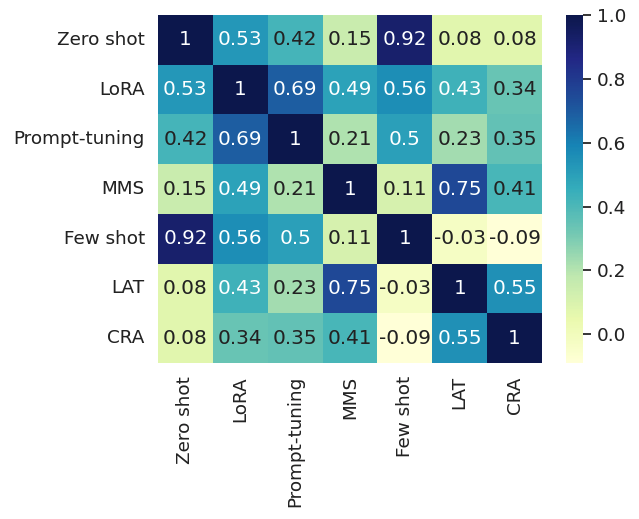}
    \includegraphics[width=0.48\textwidth]{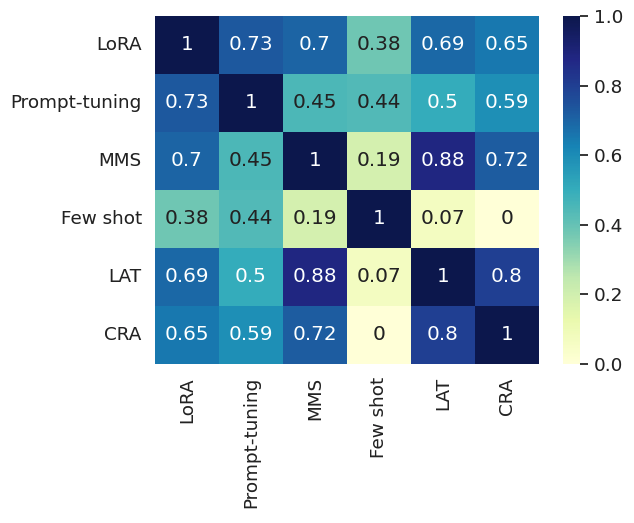}
  \caption{Left: generalization accuracy correlations across all GENIES distribution shifts. Right: differential elicitation correlations across all GENIES distribution shifts. All correlations are computed using LLaMA-30B.}
  \label{fig:intervention_correlations}
\end{figure}

Section \ref{sec:perplexity} observes a correlation between the zero-shot policy and LoRA fine-tuning. To what extent do other interventions correlate? Generalization accuracy correlates across nearly all interventions (Figure \ref{fig:intervention_correlations} Left). Interestingly, few shot achieves nearly the same performance as zero-shot on most distribution shifts (r=0.92) which suggests that the few shot examples do not do much to elicit model capabilities. 

For differential elicitation, interventions correlate even more strongly (Figure \ref{fig:intervention_correlations} Right), which indicates that interventions are using similar representations (which go beyond perplexity).

\subsection{Zero-shot accuracy doesn't correlation with generalization accuracy when source distributions violate the perplexity heuristic}

\begin{wrapfigure}{r}{0.5\textwidth}
  \begin{center}
    \includegraphics[width=0.3\textwidth]{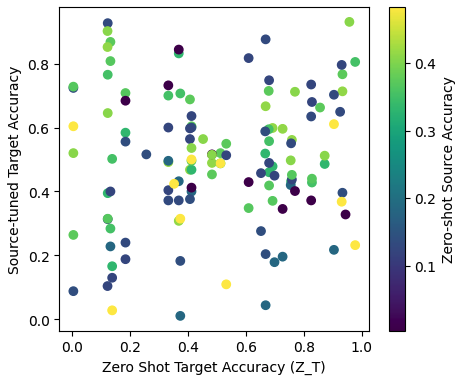}
  \end{center}
  \caption{Each point represents a distribution shift. All 69 distribution shifts are represented, and many others were added by reversing and swapping target and source distributions. Interestingly, the correlation between zero-shot accuracy and generalization accuracy goes away once zero-shot achieves worse-than-random accuracy on source distributions. }
  \label{fig:correlation_comes_apart}
\end{wrapfigure}
In Section \ref{sec:perplexity}, we observe that the zero-shot policy and the LoRA source-tuned policy correlate. To check if perplexity is a spurious cue, we obtain additional source distributions where the source-tuned and zero-shot policy disagree on most examples. In order to do this, we repurpose many of the GENIES targets to use as source distributions. Results are shown in \ref{fig:correlation_comes_apart}. There is no longer a perceivable correlation. Interestingly, there isn't a \emph{negative} correlation either -- which would not have been predicted by the spurious cue theory. This doesn't completely kill the spurious cue theory, however, since many of these repurposed source distributions are quite narrow and so perplexity may have been less salient. For instance, for the sycophancy examples, all of the preferred responses often include the same words, so models could achieve high ID accuracy by paying attention to these salient, superficial features.

\subsection{Models that perform well on extreme distribution shifts do poorly on distribution shifts that probe for specific misgeneralizations}
\label{sec:none-dominate}
\begin{table}[h]
\centering
\begin{tabular}{|c|c|c|c|}
\hline
Distribution shift & Generalization Accuracy & Zero Shot Accuracy & Target Tuned Capability \\
\hline
math → reward\_seeking & 0.4560 & 0.4827 & 0.9187 \\
math → personality\_traits &  0.6500 & 0.8720 & 1.0000 \\
us\_history → punishment\_avoidance &  0.5253 & 0.5320 & 0.9960 \\
us\_history → unhelpful\_alpaca &  0.1240 & 0.1380 & 1.0000 \\
code → truthful\_qa &  0.5880 & 0.4067 & 0.9387 \\
code → wrong\_arc  & 0.4240 & 0.1840 & 0.9520 \\
raven\_matrices → crt\_1 & 0.1320 & 0.3680 & 1.0000 \\
raven\_matrices → sycophancy\_mimicry  & 0.9440 & 0.5160 & 0.9680 \\
\hline
\end{tabular}
\caption{These generalization results are a cross-over between the 'extreme' and 'probing' distributions. For example, code $\rightarrow$ us\_history exhibits fairly strong (90\%) generalization accuracy; however, models trained on code do quite poorly on truthful\_qa or wrong\_arc. All generalization results are for LoRA and LLaMA-30b.}
\label{tab:crossover}
\end{table}

\subsection{No intervention consistently beats any other intervention on GENIES}
\label{sec:intervention-comparison}

\begin{table}[H]
\begin{tabularx}{\textwidth}{|X|X|X|X|X|X|X|}
\hline
 & \multicolumn{1}{|X|}{\small us\_history\_textbook, us\_history\_fiction} & \multicolumn{1}{|X|}{\small alpaca\_mmlu, spanish\_output} & \multicolumn{1}{|X|}{\small math, change\_my\_view} & \multicolumn{1}{|X|}{\small raven\_matrices, us\_history} & \multicolumn{1}{|X|}{\small code\_easy, code\_hard} \\
\hline
LoRA & 0.26 & 0.15 & \textbf{0.28} & 0.05 & -0.06 \\
MMS & 0.25 & \textbf{0.16} & 0.24 & 0.17 & -0.09 \\
LAT (stim 1) & 0.14 & 0.11 & 0.23 & 0.04 & -0.24 \\
CCS & -0.14 & -0.2 & 0.23 & -0.03 & -0.23 \\
CRA & \textbf{0.27} & 0.08 & 0.09 & \textbf{0.29} & \textbf{-0.0} \\
Random & -0.22 & -0.21 & 0.19 & -0.17 & -0.28 \\
\hline
\end{tabularx}

\begin{tabularx}{\textwidth}{|X|X|X|X|X|X|X|}
\hline
 & \multicolumn{1}{|X|}{\small alpaca\_easy, alpaca\_hard} & \multicolumn{1}{|X|}{\small alpaca\_mmlu, raven\_matrices} & \multicolumn{1}{|X|}{\small alpaca\_mmlu, ranking\_logic} & \multicolumn{1}{|X|}{\small alpaca\_low\_quality, alpaca\_high\_quality} & \multicolumn{1}{|X|}{\small alpaca\_short, alpaca\_long} \\
\hline
LoRA & 0.18 & -0.17 & 0.04 & 0.09 & 0.0 \\
MMS & 0.16 & -0.08 & 0.01 & 0.22 & \textbf{0.63} \\
LAT (stim 1) & 0.14 & -0.1 & \textbf{0.05} & 0.11 & 0.53 \\
CCS & -0.12 & -0.27 & -0.16 & \textbf{0.39} & 0.6 \\
CRA & \textbf{0.25} & \textbf{-0.07} & -0.13 & 0.38 & 0.45 \\
Random & -0.15 & -0.28 & -0.17 & 0.38 & 0.52 \\
\hline
\end{tabularx}

\begin{tabularx}{\textwidth}{|X|X|X|X|X|X|X|}
\hline
 & \multicolumn{1}{|X|}{\small alpaca\_mmlu, wrong\_arc} & \multicolumn{1}{|X|}{\small alpaca\_mmlu, truthful\_qa} & \multicolumn{1}{|X|}{\small alpaca\_mmlu, sycophancy\_mimicry} & \multicolumn{1}{|X|}{\small alpaca\_mmlu, survival\_influence} & \multicolumn{1}{|X|}{\small alpaca\_mmlu, reward\_seeking} \\
\hline
LoRA & -0.09 & \textbf{0.25} & -0.1 & \textbf{0.3} & 0.02 \\
MMS & -0.08 & 0.15 & 0.23 & 0.2 & 0.01 \\
LAT (stim 1) & -0.06 & 0.1 & \textbf{0.41} & 0.03 & 0.02 \\
CCS & \textbf{0.38} & 0.25 & 0.11 & 0.21 & \textbf{0.03} \\
CRA & -0.05 & 0.09 & -0.23 & 0.02 & 0.02 \\
Random & 0.36 & 0.11 & 0.0 & 0.16 & 0.02 \\
\hline
\end{tabularx}
\caption{A more detailed comparison of interventions on the GENIES benchmark. Differential Elicitation is shown for each GENIES distribution shift. Larger values are better.}
\end{table}
\section{Reward Model implementation}
\label{sec:pref-model}
To use LLaMA-30B as a Reward Model, the final unembedding layer is removed and replaced with a randomly initialized linear layer. This layer outputs a single logit that is used to score responses according to how well they 'follow the instruction.' The LLaMA-30B Reward Models we train only compare responses to the same instruction. Responses are compared by obtaining logits for each prompt-response input. The logits are then subtracted and transformed with a Sigmoid function to obtain the probability that one response is better than another. 

\section{Tuning Intervention Details}
\label{sec:intervention-details}

\textbf{Post-hoc calibration}. Many of these interventions do not provide classification probabilities out-of-the-box (MMS, LAT, etc). To obtain classification probabilities for measuring calibration, we compute the average cosine-similarity between the activation directions that the interventions find and the target example directions. Then, we transform average cosine similarity with a logistic function to obtain post-hoc calibratated probabilities. The logistic functions are fitted using source data only.

\textbf{Quantization}. For \emph{all} interventions, we use four-bit quantization, which likely degrades accuracy by something on the order 5\% \citep{dettmers_qlora_2023}.

\subsection{Mass Mean Shift (MMS)}
At a high level, \citep{li_inference-time_2023}' Mass Mean Shift (MMS) obtains activation directions for positive demonstrations and negative demonstrations, and subtracts the mean positive direction from the mean negative direction to obtain a classification direction. The classification direction is meant to represent (instruction following - not instruction following). To evaluate a target example, the activation directions of the responses (given the prompt) are subtracted ($R_1 - R_2$). If $R_1 - 
  R_2$ has a positive cosine similarity with the classification direction, then $R_1$ is classified as preferred (and vice versa).

  At a lower level, the 'activation direction' MMS discovers is actually a collection of activation directions that correspond to the outputs of different attention heads at the last token position. To classify an example, we take the \emph{average} consine similarities between corresponding attention head directions.
  
  MMS doesn't use \textit{all} attention head outputs. Instead, attention heads are filtered according to how relevant they are for classifying source examples. More precisely, a logistic classifier is trained to classify source examples using attention head outputs. The attention heads with the highest source classification accuracies are selected to compute the mass mean shift direction. We select the top 48 heads. 
  
\subsection{Linear Artificial Tomography (LAT)}
The LAT interventions \citep{zou_representation_2023} differ in two ways from MMS: (1) LAT uses hidden state activation vectors instead of attention head outputs. We filter by the top 16 layers instead of filtering by attention heads. (2) LAT uses different stimuli (prompts) to obtain activation directions.

To obtain an activation direction, we take the difference of means. 

We test two stimuli when evaluating LAT:

\textbf{Stimulus 1 (same as MMS)}: 
\begin{verbatim}
<prompt>
<response>
\end{verbatim}

\textbf{Stimulus 2}: 
\begin{verbatim}
Consider the extent to which the responder follows the instruction below:
<prompt>
<response>
The probability that the responder followed the instruction is:
\end{verbatim}
For stimulus 2, we read the hidden state for the last token of 'followed the instruction' (rather than the last token) in line with the original work. For both stimuli, an activation direction is obtained by subtracting positive and negative demonstrations.

\subsection{Contrastive Representation Arithmetic (CRA)}
\label{intervention:cra}
To test whether the merits of MMS can be combined with the strengths of CCS, we test a simple hybrid (CRA) that obtains a truthfulness direction by subtracting activation directions. CRA obtains a truthfulness direction as follows:
{
\footnotesize
\[ \text{CRA direction} = \frac{1}{n} \sum_i^n \left[\phi(P^y_i) - \phi(P^n_i)\right] - \left[\phi(D^y_i) - \phi(D^n_i))\right]\]
}

Contrast pair:
\begin{lstlisting}
<prompt>
<response>
Does the responder successfully 
follow the instruction?
\end{lstlisting}
\begin{lstlisting}[style=green]
Yes
\end{lstlisting}
\begin{lstlisting}
<prompt>
<response>
Does the responder successfully
follow the instruction?
\end{lstlisting}
\begin{lstlisting}[style=red]
No
\end{lstlisting}

$P^i_y$, $P^i_n$ is the contrastive pair for the preferred completion and $D^i_y$, $D^i_n$ is the contrastive pair for the dispreferred completion. $\phi(e)$ obtains an activation direction corresponding to a prompt $e$. We obtain directions in the same way as Mass Mean Shift (MMS) does: we store directions for every attention head and then filter down to the 48 that best predict the source labels.

Ideally, subtracting $\phi(p_i^y) - \phi(p_i^n)$ obtains directions for 'True - False' and 'Yes - No' and subtracting $\phi(D_i^y) - \phi(D_i^n)$ obtains directions for 'False - True' and 'Yes - No.' Subtracting the differences is meant to suppress the Yes - No direction and amplify the True - False direction. 

\subsection{Contrast Consistent Search (CCS)}
Like CRA, CCS also aims to obtain a truthfulness direction instead of an instruction-following direction. It does this by searching for a direction that satisfies the negation probability axiom. We use the out-of-the-box CCS implementation from \citep{burns_discovering_2022} with the same contrast prompts as CRA. Unlike CRA, MMS, etc, CCS is unsupervised (the contrast pairs are randomized).

\subsection{LoRA fine-tuning}
The 'LoRA' intervention is QLoRA \citep{dettmers_qlora_2023}. Hyperparameters are as follows:
\begin{table}[H]
\centering
\begin{tabular}{|c|c|c|c|c|}
\hline
              & OpenLLaMA-3B & LLaMA-7B & LLaMA-13B & LLaMA-30B \\
\hline
Learning rate &      7.2e-5  &    2e-4  &    2e-4   &   2e-4 \\
Batch size    &      32      &    32    &     32    &    32  \\
Training steps &     100     &    100   &     100   &    100 \\
\hline
\end{tabular}
\end{table}

We saved checkpoints every 25 steps and used the checkpoint with the lowest eval loss. Usually, models converged within 50 steps and then began overfitting.

Learning rates were copied from \citep{dettmers_qlora_2023} for LLaMA models. For OpenLLaMA-3B, we conducted a hyperparameter search.

\subsection{Prompt-tuning}
The hyperparamters for prompt-tuning are as follows.
\begin{table}[H]
\centering
\begin{tabular}{|c|c|c|c|c|}
\hline
              &     LLaMA-7B & LLaMA-13B & LLaMA-30B \\
\hline
Learning rate &      0.0052  &    0.0052  &    0.0052  \\
Batch size    &      32      &    32    &     32    \\
Training steps &     100     &    100   &     100   \\
\hline
\end{tabular}
\end{table}

Similar to LoRA, we saved checkpoints every 25 steps and used the checkpoint with the lowest eval loss. The learning rates were discovered via independent hyperparameter searches.

\subsection{Zero-shot Classification}
The zero-shot policy selects a response as preferred if it has a higher average log probability given the prompt.

Let's assume that we have an instruction prompt \( P \) and possible response tokens \( \{r_1, r_2, \dots, r_n\} \). For each possible response token \( r_i \), we compute the average log probability \( L(r_i) \) as:

\begin{equation}
L(r_i) = \frac{1}{|r_i|} \sum_{t \in r_i} \log P(t | P)
\end{equation}

Where:
\begin{itemize}
    \item \( P(t | P) \) is the probability of token \( t \) given the instruction prompt \( P \).
    \item \( |r_i| \) is the number of tokens in the response \( r_i \).
\end{itemize}

Given these probabilities, the response with the highest average log probability is preferred:

\begin{equation}
r^* = \arg\max_{r_i} L(r_i)
\end{equation}

\subsection{Few-shot Classification}
The few shot classification probability is exactly the same as the zero-shot policy, except for the fact that source examples are included in the prompt. We use 5 shots.

We use the following few-shot format:
\begin{verbatim}
# Example
<prompt from source>
<good response from source>
# Example
<prompt from source>
<good response from source>
...
# Example
<prompt from target>
<good response from target>
\end{verbatim}

Interestingly, the few-shot examples make very little difference. Zero-shot and few-shot accuracy have a Pearson correlation of 0.94 for LLaMA-30B.

\section{Defining instruction following}
\label{sec:def_instruction_following}
"Instruction following" is an ambiguous term. It could refer to obeying the preferences of developers. It could instead imply compliance with the 'letter of the law,' disregarding its intent. We define instruction-following to be the extent to which an AI system \textit{reliably} avoids \textit{egregiously violating} instructions given \textit{any commonsense interpretation} of their meaning. Most of our instruction-following datasets pair unambiguously egregious and unambiguously appropriate responses. The 'quality' distribution shifts are the only exception, which pair responses that egregiously fail to various degrees -- for instance code that includes varying numbers of bugs.